\documentclass[peerreview]{IEEEtran}
\usepackage{cite} % Tidies up citation numbers.
\usepackage{url} % Provides better formatting of URLs.
\usepackage[utf8]{inputenc} % Allows Turkish characters.
\usepackage{booktabs} % Allows the use of \toprule, \midrule and \bottomrule in tables for horizontal lines
\usepackage{graphicx}
\usepackage{subcaption}
\usepackage{graphicx}
\usepackage{times}
\usepackage{epsfig}
\usepackage{graphicx}
\usepackage{amsmath}
\usepackage{amssymb}
\usepackage{adjustbox}
 \usepackage{verbatim}
\usepackage{bm}
\usepackage{subcaption}
\usepackage{amsmath}
\usepackage{array}
\begin{document}
%\begin{titlepage}
% paper title
% can use linebreaks \\ within to get better formatting as desired
\title{Exploring the Effect of Sparse Recovery on the Quality of Image Superresolution}

% author names and affiliations

\author{Antonio Castro\\
Department of Mechanical Engineering\\
Fresno City College\\
}
\date{25/4/15}

% make the title area
\maketitle
%\tableofcontents
%\listoffigures
%\listoftables
%\end{titlepage}

%\IEEEpeerreviewmaketitle
\begin{abstract}
Dictionary learning can be used for image superresolution by learning a pair of coupled dictionaries of image patches from high-resolution and low-resolution image pairs such that the corresponding pairs share the same sparse vector when represented by the coupled dictionaries. These dictionaries then can be used to  to reconstruct the corresponding high-resolution patches from low-resolution input images based on sparse recovery. The idea is to recover the shared sparse vector using the low-resolution dictionary and then multiply it by the high-resolution dictionary to recover the corresponding high-resolution image patch. In this work, we study the effect of the sparse recovery algorithm that we use on the quality of the reconstructed images. We offer empirical experiments to search for the best sparse recovery algorithm that can be used for this purpose.

\end{abstract}

\section{Introduction}
Image super-resolution is an important problem in computer vision due to its numerous practical applications and significant impact on various domains.  Super-resolution techniques can significantly enhance the visual quality of images by increasing their resolution without improving the imaging device. This is particularly valuable in applications where high-quality images are crucial, such as medical imaging, satellite imagery, and surveillance. Higher-resolution images provide more detailed information, leading to improved accuracy and reliability in image analysis tasks, such as object detection ~\cite{wang2022remote}, recognition \cite{yang2018long}, and segmentation~\cite{sert2019new}. Super-resolution also helps to reveal finer details that might be essential for precise analysis.

Superresolution has been explored extensively in the literature and there are many existing approaches for this purpose. To date,  explorations are continued to improve the existing methods and pushing the boundaries to problems that are considered ill-posed for the moment based on suitable quality metrics for this problem~\cite{yeganeh2012objective,yeganeh2015objective,zhou2021image,fang2018blind,zhou2022quality,zhao2021learning}.
Dictionary learning is one of these methods that can be employed for image superresolution by training a pair of coupled dictionaries consisting of image patches derived from both high-resolution and low-resolution image pairs \cite{yang2008image,yang2010image,rostami2011image,rehman2012ssim,wang2015deep}. The objective is to ensure that the corresponding pairs of image patches share the same sparse vector representation. These dictionaries are subsequently utilized to reconstruct the corresponding high-resolution patches from low-resolution input images using a two stage process. First, a sparse recovery technique is used to reconstruct the corresponding shared sparse vector using the low-resolution dictionary. The shared sparse vector then used to reconstruct the high-resolution image by multiplying it by the high-resolution dictionary. This approach has been shown to be quite effective in various problem~\cite{yang2011multitask,xie2014single,zhang2011efficient,ayas2020single,ahmed2016single}.

The dictionary learning process begins by collecting a dataset comprising pairs of high-resolution images and their corresponding low-resolution counterparts. We can prepare such a dataset using  crawling algorithms from the internet and limit our dataset to include those images that are relevant to our particular application of interest. In most cases, the low-resolution images are simply generated from the high-resolution images. Next, the high-resolution images are divided into small, overlapping patches. Similarly, the low-resolution images are subjected to patch extraction. A training set is then created by pairing the high-resolution patches with their corresponding low-resolution patches.
Next, dictionary learning algorithms, such as K-SVD ~\cite{aharon2006k} are applied to learn a pair of dictionaries—one for the high-resolution patches and another for the low-resolution patches. These dictionaries are trained in a coupled manner to capture the underlying structures and textures present in both high-resolution and low-resolution images, ensuring that the corresponding pairs share the same sparse vector representation.

To reconstruct a high-resolution image from a low-resolution version, we rely on  sparse recovery algorithms~\cite{mohimani2008fast,alamdari2023modified,zhang2011sparse,fu2014block,marques2018review,becker2011nesta}. When a   low-resolution input image is provided, it is first divided into patches. For each low-resolution patch, the best-matching high-resolution patches from the learned dictionaries are determined, utilizing the shared sparse vector representation by recovering it using the selected sparse recovery algorithm. The selection of the sparse recovery is a design choice and can affect the eventual quality. The high-resolution patches are then reconstructed by leveraging the shared sparse recovery algorithm and multiplying it by the high-resolution dictionary.

The sparse recovery algorithm that we use in the reconstruction phase affects the quality of the generated high-resolution image. It a natural question what algorithm works better for image superresolution.
To assess the effect of sparse recovery algorithms on the quality of the reconstructed images, we conduct empirical experiments where we compare and evaluate different sparse recovery algorithms for image superresolution. These experiments involve applying each algorithm to the superresolution task and measuring the resulting image quality metrics on a set of test images such as peak signal-to-noise ratio (PSNR). By systematically examining the performance of various sparse recovery algorithms, we aim to identify the most effective and suitable algorithm for achieving high-quality reconstructions in image superresolution tasks.
Through our investigation, we contribute insights into the impact of the sparse recovery algorithm on the overall quality of reconstructed images in the context of dictionary learning-based image superresolution. Our empirical experiments provide valuable evidence to guide the selection of the most appropriate sparse recovery algorithm for this specific purpose, improving the effectiveness and performance of image superresolution techniques.

 \section{Background and Related Work}

 \subsection{Classic Sparse Recovery}
Sparse recovery is a powerful method employed to reconstruct signals-of-interest from incomplete measurements that fall below the critical threshold of Nyquist rate \cite{4,qq4}. This innovative approach has found diverse applications in image processing   and computer vision  application \cite{re3,re5,rostami2012gradient,re9,rostami2015surface,re4,re6,re7,re8,rostami2015surface,rostami2013compressed,rostami2012image,hashemi2016efficient,rostami2012gradient,wisdom2017building,marques2018review,marques2019deep,xie2019fast,rostami2013compressed,song2019coupled}, effectively addressing many challenging problems in these fields.
 Sparse recovery   leverages sparsity as prior information about the source signal. A study conducted by \cite{re4} demonstrates how incorporating sparsity improves the robustness of signal recovery against outliers and noise.  

At a fundamental level, CS exploits sparsity as a prior on the signal to be estimated and recovers an approximation to the signal as a unique solution to a convex optimization problem. Let $\textbf{y} \in \mathbb{R}^n$ be a source signal which has a sparse representation with respect to some discrete basis $D\in\mathbb{R}^{n\times m}$, that is $\textbf{x}=D\alpha$, where $\\alpha$ is a sparse signal with $\|\alpha\|_0=k, \, k<n$\footnote{Here $\|\cdot\|_0$ denotes the $\ell_0$-norm that counts the number of non-zero elements of $\alpha$.}. One of the central results of the theory of CS states that a $k$--sparse signal $\alpha$ can be reliably recovered from $m>k \log (n/k)$ measurements acquired according to
\begin{equation}  
\bm{x} = D \alpha+\textbf{n},
\label{eq2}
\end{equation}
where $\alpha \in \mathbb{R}^m, n<m$    \cite{4,qq4}. Here, $D \in \mathbb{R}^{m\times n}$ is a full-rank matrix known as the {\em sensing matrix} and $\textbf{n}$ denotes measurement noise, which is usually assumed to be additive Gaussian noise. Various approaches have been developed to recover the sparse vector $\textbf{x}$ in Eq.~\eqref{eq2}. We briefly mention a few popular methods.
Basis Pursuit (BP) ~\cite{chen1994basis} is a specific type of $\ell_1$-norm minimization that seeks the sparsest solution to an underdetermined linear system. It finds the minimum $\ell_1$ norm solution that satisfies the linear equations ~\eqref{eq2}. Orthogonal Matching Pursuit (OMP)~\cite{tropp2007signal,wang2012generalized} is an iterative greedy algorithm used for sparse recovery. It starts with an empty set of selected indices and iteratively adds the index that maximizes the correlation with the current residual until a stopping criterion is met.  
Iterative Hard Thresholding (IHT) ~\cite{blumensath2009iterative}  is an iterative algorithm that performs hard thresholding in each iteration, which involves zeroing out small entries in the solution to promote sparsity. The Iterative Soft-Thresholding Algorithm (ISTA) \cite{beck2009fast} is an improvement over IHT based on the idea of soft thresholding. Quadratic Programming (QP) also has been used for this purpose to improve the convergence speed \cite{li2013sparse}.
Least Absolute Shrinkage and Selection Operator (LASSO)~\cite{ranstam2018lasso,roth2004generalized} is a regression method that adds an $\ell_1$-norm penalty to the objective function, encouraging sparsity in the solution. The Bergman-based method~\cite{goldstein2009split} relies on splitting optimization methods \cite{annergren2012admm,hao2016testing} for more efficient solution recovery. SL0 algorithm~\cite{mohimani2008fast} offers a solution based on minimizing the $\ell_0$-norm. To this end, $\ell_0$-norm is approximated using a smooth differentiable function in order to benefit from gradient-based optimization methods.

  \subsection{Dictionary Learning}
A major question for sparse recovery is how to find the dictionary $D$ in Eq.~\eqref{eq2}. Dictionary learning is a powerful technique where the goal is to find an overcomplete dictionary, denoted as $\mathbf{D} \in \mathbb{R}^{n \times m}$. To solve for the dictionary matrix, we rely on a set of training data. Given a set of training data samples ${\mathbf{x}_1, \mathbf{x}_2, ..., \mathbf{x}_N}$, dictionary learning seeks to find $\mathbf{D}$ and the corresponding sparse codes ${\alpha_1, \alpha_2, ..., \alpha_N}$ such that each data sample $\mathbf{x}_i$ can be approximated as a sparse linear combination of the dictionary atoms, i.e., $\mathbf{x}_i \approx \mathbf{D}\alpha_i$, where $\alpha_i$ is a sparse vector with only a few non-zero entries.
The dictionary learning problem can be formulated as an optimization problem, aiming to minimize the representation error while enforcing sparsity on the sparse codes. A common approach is to solve the following optimization problem:
\begin{equation}
\min_{\mathbf{D}, {\alpha_i}} \sum_{i=1}^{N} \|\mathbf{x}_i - \mathbf{D}\alpha_i\|_2^2 + \lambda \sum_{i=1}^{N} \|\alpha_i\|_0,
\label{eq1}
\end{equation}
where the first term represents the reconstruction error, measuring the fidelity of the approximations, and the second term promotes sparsity by imposing an $\ell_0$-norm penalty on the sparse codes with a regularization parameter $\lambda$. Solving this optimization problem iteratively, often using methods like K-SVD, enables learning a dictionary that can effectively represent the given data in a sparse and compact manner. The common approach generally involves an alternating scheme where at iterative steps, the dictionary is fixed and the sparse representations are updated and the reverse is preformed to update the dictionary. The resulting learned dictionary and sparse codes can be employed in various applications, such as denoising, compression, and feature extraction, achieving superior performance compared to fixed and predefined dictionaries.
In this study, we are interested in an extension of the original dictionary learning problem in Eq.~\eqref{eq1}, called coupled dictionary learning for image superresolution.

  \section{Coupled Dictionary Learning for Image Superresolution}

Image superresolution is a crucial task within the domain of computer vision, with the primary objective of augmenting the resolution of a low-resolution (LR) image to generate its high-resolution (HR) counterpart. Conventional interpolation-based methods often result in suboptimal outcomes, characterized by blurriness and unrealistic appearance~\cite{dengwen2010edge,han2013comparison}. To overcome these limitations, coupled dictionary learning has emerged as a highly promising approach for image superresolution, exhibiting enhanced performance while preserving intricate details~\cite{yang2008image,yang2010image,wang2015deep}. By capitalizing on the inherent relationship between LR and HR image patches, coupled dictionary learning efficiently captures their shared characteristics, leading to significantly improved HR reconstructions. Notably, this approach excels in accurately reconstructing texture details and sharp edges, making it a valuable tool in various image processing and computer vision applications. The use of coupled dictionary learning enables the generation of visually appealing and high-quality HR images, surpassing the capabilities of traditional interpolation techniques.

The basic idea of coupled dictionary learning is to learn a pair of dictionaries, one for LR patches and another for HR patches, that are jointly optimized to enforce consistency between the LR and HR domains. Let's denote the LR dictionary as $\mathbf{D}_{LR} \in \mathbb{R}^{n_{\text{LR}} \times m}$ and the HR dictionary as $\mathbf{D}_{HR} \in \mathbb{R}^{n_{\text{HR}} \times m}$, where $n_{\text{LR}}$ and $n_{\text{HR}}$ are the dimensions of LR and HR patch space, respectively, and $m$ is the number of atoms in each dictionary. Given a LR image $\mathbf{Y}$, the goal is to find an HR estimate $\mathbf{X}$ by jointly inferring the sparse codes $\alpha_{LR}$ and $\alpha_{HR}$ representing the LR and HR patches, respectively.
The coupled dictionary learning problem can be formulated as follows:
\begin{equation}
\min_{\mathbf{D}_{LR}, \mathbf{D}_{HR}, \alpha_{LR}, \alpha_{HR}} \sum_{i=1}^{N} \left\|\mathbf{x}_i - \mathbf{D}_{LR}\alpha_{\text{LR}, i}\right\|_2^2 + \lambda_{\text{LR}} \left\|\alpha_{\text{LR}, i}\right\|_1 + \left\|\mathbf{y}_i - \mathbf{D}_{HR}\alpha_{\text{HR}, i}\right\|_2^2 + \lambda_{\text{HR}} \left\|\alpha_{\text{HR}, i}\right\|_1,
\end{equation}
where $\mathbf{y}_i$ and $\mathbf{x}_i$ are the LR and HR patches extracted from $\mathbf{Y}$ and the ground-truth HR image $\mathbf{X}$, respectively. The terms $\left\|\mathbf{y}_i - \mathbf{D}_{LR}\alpha_{\text{LR}, i}\right\|_2^2$ and $\left\|\mathbf{x}_i - \mathbf{D}_{HR}\alpha_{\text{HR}, i}\right\|_2^2$ measure the reconstruction errors for LR and HR patches, while the $\ell_1$-norm regularization terms $\lambda_{\text{LR}} \left\|\alpha_{\text{LR}, i}\right\|_1$ and $\lambda_{\text{HR}} \left\|\alpha_{\text{HR}, i}\right\|_1$ encourage sparsity in the sparse codes, promoting more accurate and compact representations.

The solution of the coupled dictionary learning problem is accomplished through a series of iterative updates, refining both the dictionaries and sparse codes until convergence is achieved. To accomplish this, a diverse range of optimization algorithms has been employed, tailored to address the intricacies of the task. Among these algorithms, the Alternating Direction Method of Multipliers (ADMM) and Online Dictionary Learning have emerged as prominent choices, offering efficient and effective solutions.
During the iterative process, ADMM and Online Dictionary Learning algorithms iteratively refine the LR and HR dictionaries, $\mathbf{D}_{LR}$ and $\mathbf{D}_{HR}$, respectively, as well as the associated sparse codes $\alpha_{LR}$ and $\alpha_{HR}$. By iteratively updating these components, the coupled dictionary learning procedure converges to dictionaries that accurately represent the interdependencies between LR and HR image patches, leading to more precise and robust HR reconstructions.
Upon the completion of the coupled dictionary learning process, the acquired HR estimate, denoted as $\mathbf{X}$, can be synthesized. This is achieved by utilizing the inferred sparse codes $\alpha_{HR}$ along with the HR dictionary $\mathbf{D}_{HR}$. Leveraging these learned components, the final HR estimate $\mathbf{X}$ showcases a significant improvement in resolution and detail, outperforming traditional interpolation-based techniques.
The integration of such optimization algorithms within coupled dictionary learning ensures that the dictionaries and sparse codes effectively capture the underlying structures and relationships between LR and HR image patches. Consequently, the synthesized HR estimate $\mathbf{X}$ emerges as a highly refined representation, replete with enhanced textures, sharp edges, and finer visual nuances.

Coupled dictionary learning has exhibited promising outcomes not only in image superresolution tasks but also in other machine learning (ML) problems, showcasing its versatility and effectiveness across diverse domains. One notable application of coupled dictionary learning is its utilization in scenarios where the objective is to couple two vector spaces with differing sizes. For instance, Li et al. \cite{lu2017simultaneous} and Wang et al. \cite{wang2020coupled} harnessed dictionary learning to enhance face recognition performance, showcasing its potential in this critical area. Furthermore, researchers have leveraged coupled dictionary learning for image annotation tasks, as demonstrated in the work of Roostaiyan et al. \cite{roostaiyan2022toward}, further highlighting the technique's applicability in computer vision tasks. The success of coupled dictionary learning extends to zero-shot learning in visual recognition \cite{kolouri2018joint,rostami2022zero,kolouri2021attribute,liu2014semi,li2023hierarchical,zhu2016coupled,zhuang2013supervised,li2021zero} and reinforcement learning \cite{rostami2020using,isele2016using,rostami2017multi} domains, where it has been employed to enable learning without direct training data or to enhance policy optimization. Moreover, the research conducted by Jegou et al. \cite{ji2019partial} delved into the application of coupled dictionary learning for image clustering, showcasing its capacity to handle data grouping challenges. These examples collectively illustrate the broad spectrum of problems to which coupled dictionary learning has been applied, spanning from image classification to brain image analysis, highlighting its potential as a valuable tool in various ML applications.

In our research project, our main focus revolves around investigating the impact of different sparse recovery algorithms on the quality of superresolution images obtained using trained coupled dictionaries. Specifically, we are interested in understanding how the choice of sparse recovery algorithm influences the final output of the superresolution process. To put it simply, we treat the sparse recovery algorithm as a critical hyperparameter in the context of superresolution, which is based on coupled dictionary learning.
To achieve our research objectives, we  conduct an empirical study. This study involves   comparing and fine-tuning various sparse recovery algorithms to determine which one results in the best performance outcomes and produces superresolution images of the highest quality. By analyzing the differences in the output images, we can gain valuable insights into the strengths and weaknesses of each algorithm and identify the one that offers the most promising results.

\section{Experimental Exploration}

In our investigation, our   focus lies in examining the impact of four distinct sparse recovery algorithms on the overall quality of image superresolution. We employ a set of five commonly used images, which have gained wide recognition in  evaluating image processing algorithms.  
To achieve a comprehensive understanding of the results, we will assess the quality of the superresolved images using both quantitative and qualitative analyses.  
By expanding our scope to encompass multiple images and algorithms, we seek to provide   insights into the practical applicability and limitations of sparse recovery techniques in image superresolution.  

\subsection{Experimental Setup}
We first explain about our setup and then provide our empirical results.

\paragraph{Data:} in our experiments we rely on five images. These images are Lena, Boat, Cameraman, Butterfly, and Barbara. The images include sharp edges and texture in addition to a significant amount of details which makes them a good test-bed for image superresolution.
At the same time, they form a  diverse set   that presents varying characteristics and complexities to make our conclusions robust.  
\paragraph{Sparse Recovery Methods:}
We perform experiments
The selected sparse recovery algorithms are at the core of our research, as they have shown promise in handling the challenges of image superresolution. We have used OMP~\cite{tropp2007signal,wang2012generalized}, SL0~\cite{mohimani2008fast}, QP \cite{li2013sparse}, ISTA \cite{beck2009fast} methods in our experiments to include methods that are significantly different from each to use a representative subset of possible existing methods.

\paragraph{Evaluation} We use the Peak Signal-to-Noise Ratio (PSNR) as an objective metric  to evaluate the quality of the image reconstruction  compared to the original image. PSNR quantifies the similarity between the original image and the reconstructed  image by calculating the ratio of the peak signal power to the mean squared error (MSE) between the two images.
The PSNR is expressed in decibels (dB) and is defined as:

\[ \text{PSNR} = 10 \cdot \log_{10}\left(\frac{{\text{MAX}^2}}{{\text{MSE}}}\right) \]

where the maximum possible pixel value is represented by \text{MAX}, and the mean squared error is represented by \text{MSE}. The formula uses the standard mathematical notation for logarithms with base 10, represented by $\log_{10}$.

\subsection{Results}

In Table~\ref{tab}, we have showcased our comprehensive quantitative findings. One   observation from the results is the significant impact that the choice of sparse recovery algorithm can have on image superresolution. For instance, when comparing the outcomes for the image ``Lena," we notice a substantial difference of approximately 2.5 dB between the performances of the QP and SL0 algorithms. We can conclude that as a hyperparameter, it is important to select the best sparse recovery algorithm for optimal image superresolution results.

Furthermore, a broader analysis of the overall outcomes reveals a noteworthy trend. On average, the QP algorithm consistently emerges as the top performer among the selected methods. This compelling consistency across multiple images suggests that QP stands out as a superior choice for sparse recovery in the context of image superresolution.
Consequently, based on our comprehensive evaluation, we can   conclude that, on average, the QP algorithm represents the most favorable option for achieving high-quality image superresolution. However, it is essential to remain mindful of the specific characteristics of individual images, as certain cases may warrant consideration of other algorithms, given their varying strengths and weaknesses. Nevertheless, our quantitative results lend   support to the preference for QP as a reliable choice for effective sparse recovery in image superresolution applications.

\begin{figure}
\Large
  \centering
  \begin{subfigure}{0.18\linewidth}
    \includegraphics[width=\linewidth]{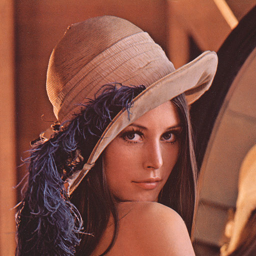}
    \caption{Lena}
  \end{subfigure}
  \begin{subfigure}{0.18\linewidth}
    \includegraphics[width=\linewidth]{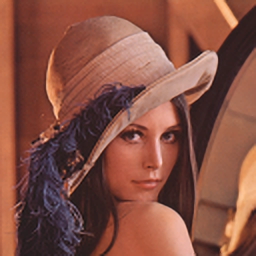}
    \caption{QP}
  \end{subfigure}
  \begin{subfigure}{0.18\linewidth}
    \includegraphics[width=\linewidth]{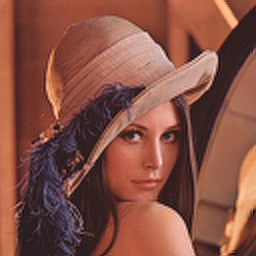}
    \caption{SL0}
  \end{subfigure}
  \begin{subfigure}{0.18\linewidth}
    \includegraphics[width=\linewidth]{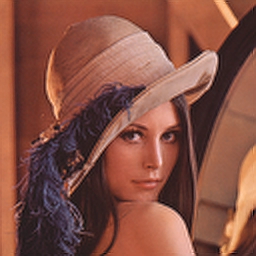}
    \caption{OMP}
  \end{subfigure}
  \begin{subfigure}{0.18\linewidth}
    \includegraphics[width=\linewidth]{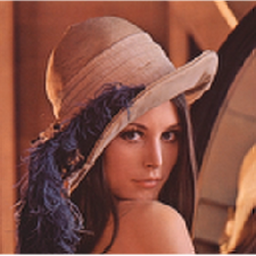}
    \caption{ISTA}
  \end{subfigure}
  
  \begin{subfigure}{0.18\linewidth}
    \includegraphics[width=\linewidth]{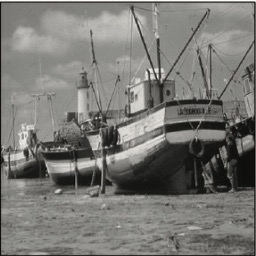}
    \caption{Boat} 
  \end{subfigure}
  \begin{subfigure}{0.18\linewidth}
    \includegraphics[width=\linewidth]{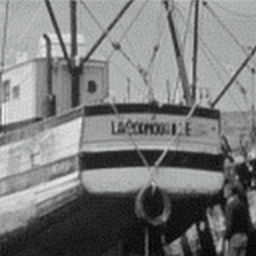}
    \caption{QP} 
  \end{subfigure}
  \begin{subfigure}{0.18\linewidth}
    \includegraphics[width=\linewidth]{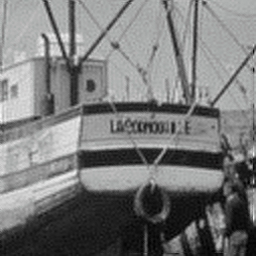}
    \caption{SL0} 
  \end{subfigure}
  \begin{subfigure}{0.18\linewidth}
    \includegraphics[width=\linewidth]{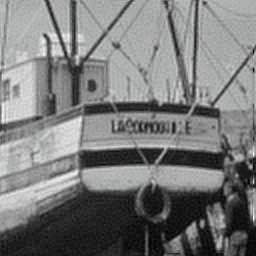}
    \caption{OMP} 
  \end{subfigure}
  \begin{subfigure}{0.18\linewidth}
    \includegraphics[width=\linewidth]{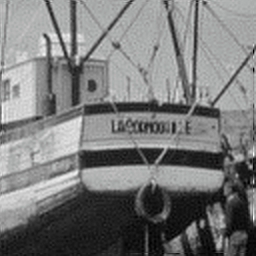}
    \caption{ISTA}
  \end{subfigure}
  
  \begin{subfigure}{0.18\linewidth}
    \includegraphics[width=\linewidth]{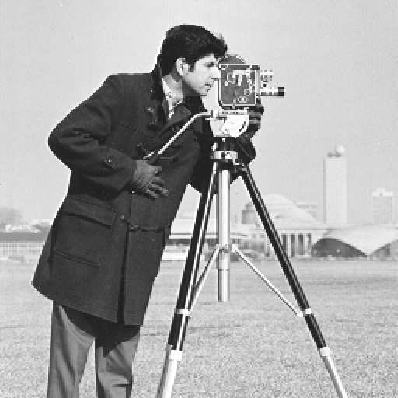}
    \caption{Cameraman}
  \end{subfigure}
  \begin{subfigure}{0.18\linewidth}
    \includegraphics[width=\linewidth]{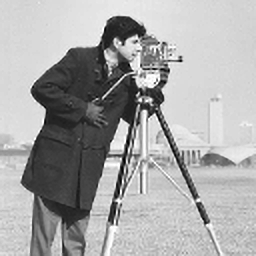}
    \caption{QP}
  \end{subfigure}
  \begin{subfigure}{0.18\linewidth}
    \includegraphics[width=\linewidth]{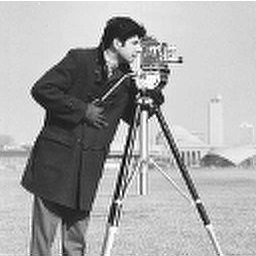}
    \caption{SL0}
  \end{subfigure}
  \begin{subfigure}{0.18\linewidth}
    \includegraphics[width=\linewidth]{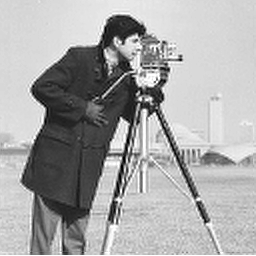}
    \caption{OMP}
  \end{subfigure}
  \begin{subfigure}{0.18\linewidth}
    \includegraphics[width=\linewidth]{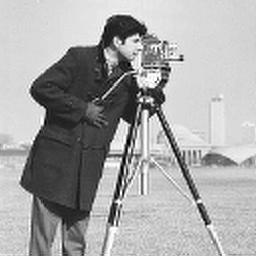}
    \caption{ISTA}
  \end{subfigure}
   \begin{subfigure}{0.18\linewidth}
    \includegraphics[width=\linewidth]{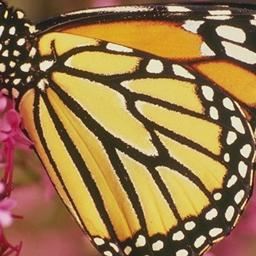}
    \caption{Butterfly}
  \end{subfigure}
  \begin{subfigure}{0.18\linewidth}
    \includegraphics[width=\linewidth]{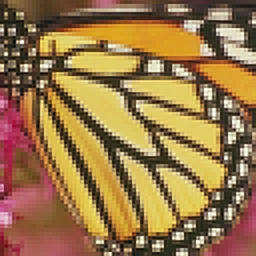}
    \caption{QP}
  \end{subfigure}
  \begin{subfigure}{0.18\linewidth}
    \includegraphics[width=\linewidth]{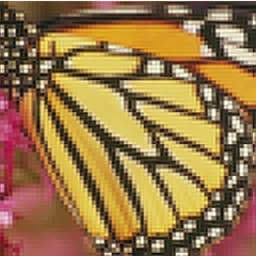}
    \caption{SL0}
  \end{subfigure}
  \begin{subfigure}{0.18\linewidth}
    \includegraphics[width=\linewidth]{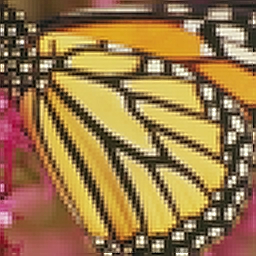}
    \caption{OMP}
  \end{subfigure}
  \begin{subfigure}{0.18\linewidth}
    \includegraphics[width=\linewidth]{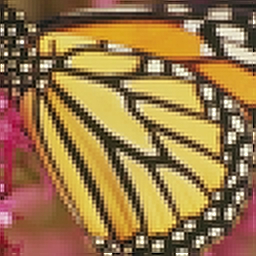}
    \caption{ISTA}
  \end{subfigure}
   \begin{subfigure}{0.18\linewidth}
    \includegraphics[width=\linewidth]{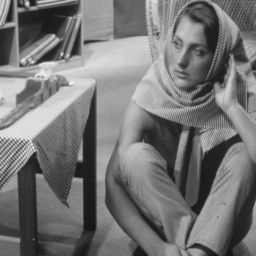}
    \caption{Barbra}
  \end{subfigure}
  \begin{subfigure}{0.18\linewidth}
    \includegraphics[width=\linewidth]{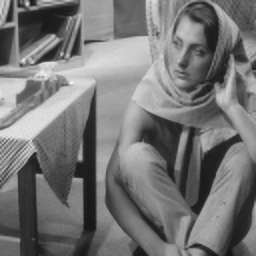}
    \caption{QP}
  \end{subfigure}
  \begin{subfigure}{0.18\linewidth}
    \includegraphics[width=\linewidth]{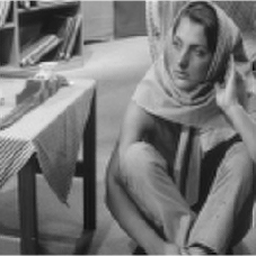}
    \caption{SL0}
  \end{subfigure}
  \begin{subfigure}{0.18\linewidth}
    \includegraphics[width=\linewidth]{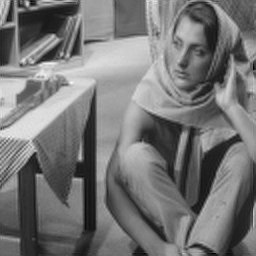}
    \caption{OMP}
  \end{subfigure}
  \begin{subfigure}{0.18\linewidth}
    \includegraphics[width=\linewidth]{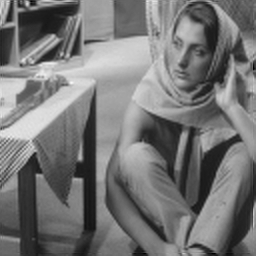}
    \caption{ISTA}
  \end{subfigure}
  \caption{Reconstructed images  using four sparse recovery methods. In each row, we have results for one of the images. The rows from left to write, correspond to ground-truth images, QP, SL0, OMP, and ISTA reconstructions.}
  \label{fig}
\end{figure}

To provide the possibility of quantitative comparison, we have also visualized our results in Figure~\ref{fig}. By a close inspection, we conclude that our quantitative results match the qualitative inspection on Figure~\ref{fig}.

\begin{table}
\large
\centering
 \begin{tabular}{|c|c|c|c|c|c|c|}
\hline
Method & Lena & Boat & Cameraman & Butterfly & Barbara & Average \\
\hline
QP & 35.04   & 12.07   & 11.81   & 19.62   & 33.81   & 22.47   \\
\hline
SL0 & 32.53  & 12.06   & 11.82   & 19.58   & 32.08   & 21.62   \\
\hline
OMP & 32.90   & 12.06   & 11.82   & 19.56   & 32.37   & 21.74   \\
\hline
ISTA & 33.71   & 12.07   & 11.82   & 19.63   & 33.02   & 22.05   \\
\hline
\end{tabular}
\caption{Performance results in PSNR for image superresolution based on dictionary learning using four sparse recovery methods. We have reported performance on five images and the average across these images.}
\label{tab}
\end{table}

\section{Conclusions and Discussions}

We conducted a comprehensive study to investigate the impact of sparse recovery algorithms on the performance of our image superresolution method based on coupled dictionary learning. Sparse recovery methods can serve as a form of regularization in the superresolution process. They help in denoising and suppressing artifacts, which are common when attempting to reconstruct high-frequency components from low-resolution input. Through a series of experiments, we aimed to understand how different sparse recovery techniques influence the quality of the superresolved images. Sparse recovery techniques excel in capturing sharp edges, fine details, and textures in images. This is crucial in superresolution, where the goal is to enhance these features and produce a visually pleasing HR result.
Our experimental results clearly demonstrate that the choice of sparse recovery algorithm does indeed play a critical role in determining the overall performance of the image superresolution process. Among the various sparse recovery algorithms we tested, one particular method stood out and consistently produced the best average performance in terms of image quality and resolution enhancement.
While our current study highlights the effectiveness of this specific sparse recovery approach, we acknowledge that there are other existing methods that we have not explored yet. Therefore, future work in this domain will involve expanding our exploration to include other sparse recovery algorithms. By systematically testing and evaluating different approaches, we aim to identify the optimal sparse recovery method that suits our specific application of interest, which is image superresolution based on coupled dictionary learning.
Ultimately, this research endeavor will enable us to refine and improve our image superresolution algorithm, leading to better results and enhanced visual quality in the superresolved images. By making well-informed choices regarding the sparse recovery approach, we can further elevate the performance and applicability of our method in practical scenarios that require high-quality superresolved images.
A limitation of our work is that we primarily focused on classic sparse recovery algorithms. More recent algorithms benefit from using deep learning in image superresolution \cite{zhang2020image,wang2015deep} and tend to lead to a better quality because these models are known to be more in line with the way that the nervous system works~\cite{selverston1976stomatogastric,morgenstern2014properties,molaie2014artificial,yuste2015neuron,sadollah2018dynamic}.
Future work includes expanding our exploration to methods based on deep learning. Another direction we see is to include more classic sparse recovery methods\cite{blumensath2009iterative,tropp2007signal,wang2012generalized,ranstam2018lasso,roth2004generalized,goldstein2009split} to make our exploration more complete.
Finally, some methods
    incorporate Noise Models  that explicitly account for noise in the low-resolution observations, resulting in more accurate and robust superresolution. Additionally, some methods  adaptively adjust the regularization parameters based on the noise characteristics, enhancing the balance between superresolution and denoising. In this work we did not explore these advanced methods. More thorough exploration should inlcude these methods.

\bibliographystyle{unsrt}
\bibliography{ref}

\end{document}